# DESIGN OF AN BASIS-PROJECTED LAYER FOR SPARSE DATASETS IN DEEP LEARNING TRAINING USING GC-MS SPECTRA AS A CASE STUDY

**Yu-Tang Chang, Shih-Fang Chen**[*]

Department of Biomechatronics Engineering, National Taiwan University,
No. 1, Sec. 4, Roosevelt Rd., Taipei, Taiwan
*Corresponding Author-- Voice: +886-2-3366-5354, Email:  sfchen@ntu.edu.tw

**Abstract:**

Deep learning (DL) models encompass millions or even billions of parameters and learn complex patterns from big data. However, not all data are initially stored in a suitable formation to effectively train a DL model, e.g., gas chromatography-mass spectrometry (GC-MS) spectra and DNA sequence. These datasets commonly contain many zero values, and the sparse data formation causes difficulties in optimizing DL models. A DL module called the basis-projected layer (BPL) was proposed to mitigate the issue by transforming the sparse data into a dense representation. The transformed data is expected to facilitate the gradient calculation and finetuned process in a DL training process. The dataset, example of a sparse dataset, contained 362 specialty coffee odorant spectra detected from GC-MS. The BPL layer was placed at the beginning of the DL model. The tunable parameters in the layer were learnable projected axes that were the bases of a new representation space. The layer rotated these bases when its parameters were updated. When the number of the bases was the same as the original dimension, the increasing percentage of the F1 scores was 8.56%. Furthermore, when the number was set as 768 (the original dimension was 490), the increasing percentage of the F1 score was 11.49%. The layer not only maintained the model performance and even constructed a better representation space in analyzing sparse datasets.

# INTRODUCTION

Deep learning (DL) models learn patterns and high-level understandings from big data and have been widely applied in various fields, e.g., computer vision and natural language processing. DL models encompass tremendous amounts of parameters and non-linear activation functions that could sketch the geometry of the data. A sub-module in DL models is usually designed to extract features from a specific data formation; the expressive power of the sub-module was not equivalent when facing different data formation. To utilize the expressive power of these sub-modules efficiently, a tailored data representation is essential.

In practical cases, data may not be stored in a representative formation and forms a "sparse" dataset. Data sparsity is categorized into two types: location-sparse and pattern-sparse. Location-sparse data means many empty sub-data points are in the whole range of data, such as 3D point cloud; pattern-sparse data represents a large number in a data "element" is zero, such as gas chromatography-mass spectrometry (GC-MS) spectrum. The "element" indicates the fundamental chunk that stored a meaningful data pattern, e.g., pixels in an image and mass spectra in a GC-MS graph. In a DL model optimization process, the location-sparsity affects the efficiency of the backpropagation learning process and could be improved by tailored data formation (Hu et al., 2019). However, the pattern-sparsity sometimes misleads the model to sketch an incorrect data geometry. Therefore, the primary purpose of this study is to explore how to properly develop a "dense" representation and reform the pattern-sparse data.

Dimension reduction was the most commonly applied technique to deal with pattern-sparsity data (Clark & Provost, 2019). It could be implemented through matrix factorization methods in linear algebra, including singular value decomposition (SVD), eigen-decomposition, non-negative matrix factorization (NMF), etc. Additionally, the dimension reduction was learnable by an autoencoder-based model. The latent vector extracted from the encoder model was also a dimension reduction method of the original data. However, the alteration of the data dimensionality was not bijective in most cases. The properties of the data space were lost, resulting in difficulties in causal discovery and model analysis.

This study proposed a DL module called the basis-projected layer (BPL) to better preserve the property of the original data space. The layer design reformatted the pattern-sparse data into a high-dimensional sphere space and kept the data geometry. The layer was evaluated on a practical sparse dataset composed of GC-MS spectra and odor labeling. The effectiveness of the proposed module was determined by the performance of the developed odor predictive models.

# MATERIALS AND METHODS

## Sparse Dataset and Model Evaluation

Coffee contains hundreds of aroma compounds resulting in its attractive odors. GC-MS provided a low detection threshold and precise determination to analyze the substances from a mixture so that it was widely applied to odor discrimination. In this study, the dataset contained 362 specialty coffee fragrance and aroma GC-MS spectra. A GC-MS spectrum was a two-dimensional matrix composed of 3375 retention time and 490 mass-to-charge ratios (m/z). The sparsity ratio of the whole GC-MS spectra in the dataset was 80.38%. There were 12 descriptor categories from the cuppers' reports to denote the coffee odor. One sample usually contained more than one odor; thus, the annotation of the coffee dataset was defined as a multi-label classification. The performance of the predictive model was mainly evaluated by the F1 score.

**Basis-Projected Layer**

The BPL was a DL module to reformat data into a sphere space:

$$Y_n = \frac{X \cdot B'_n}{\|B'_n\|^p} B'; \quad B'_n = \frac{B_n}{max(1, \|B_n\|^p)} \tag{1}$$

where a new representation ($Y \in \mathbb{R}^{(*,...,N)}$) was calculated through tensor projection over data ($X \in \mathbb{R}^{(*,...,F)}$) and the learnable bases ($B \in \mathbb{R}^{N \times F}$); the norm type (*p*) of the projected bases was set to two to acquire the Frobenius norm of the bases; *n* was the indexing of the projected bases. *F* was the dimension of a data element, and *N* was the number of learnable bases. The core concept of the BPL was a pattern-matching process between data and the learnable bases. The outcome of the layer remained zero only if the basis was orthogonal to the data element. The condition hardly occurred in practice; therefore, the sparsity would drastically be decreased to zero. Furthermore, the bases exploited a *p*-norm division to restrict the maximum length of bases. The operation maintained the numerical range of the model parameters even if the regularization of the optimizer was not set. The layer would rotate the bases of the *N*-sphere continuously to find a better space to representing pattern-sparse data in the optimization process.

**Parameter Initialization of the Basis-projected Layer**

The initial parameter method of a DL sub-module could sometimes primarily affect the final model performance (He et al., 2015). Therefore, finding an appropriate initializing method for the proposed module is crucial. Commonly, the parameters in a DL module were initialized by a Gaussian distribution and scaled up by numerical properties, e.g., the norm, and the size of the parameter. In this study, the default initializing method was based on the von Mises distribution to adequately satisfy the core concept of mapping pattern-sparse data into an *N*-sphere. Given an arbitrary center vector, a von Mises distribution sampler generated a set of bases that the angles between the center vector were a Gaussian distribution. "Concentration" was equivalent to the "variance" in the Gaussian distribution. It controlled whether the sampled bases concentrated closer to the center vector or not. The "concentration" of the von Mises distribution sampler was set to $\pi$, and therefore, directions of the initialized bases would be distributed throughout the hemisphere of an *N*-sphere space. In addition to initializing the parameter through different probability distributions, the bases could be acquired from matrix factorization methods, e.g., SVD and NMF.

**Structure of the DL Models**

The data type of the GC-MS spectrum is not commonly seen in DL applications. It is unclear whether the DL structure is applicable to extracting key features of GC-MS spectra. Voicegram is the most similar configuration compared to the GC-MS spectra. The basic DL module across speech tasks was the one-dimensional (1-d) convolutional layer. Referring to that, it was chosen as the basic module for the GC-MS data for our experiment. In this study, the structure of the odor classifier was composed of 17 layers of 1d-convolutional neural network (CNN) with residual connection (Fig. 1; He et al., 2016). The BPL was placed at the very beginning of the main CNN architecture. The Adam optimizer (Kingma & Ba, 2015) was adopted to train all models with 100k times update with a cosine annealing learning rate scheduler. The odor classifiers all reached an F1 score of one (100% accuracy) at the end of the training process.

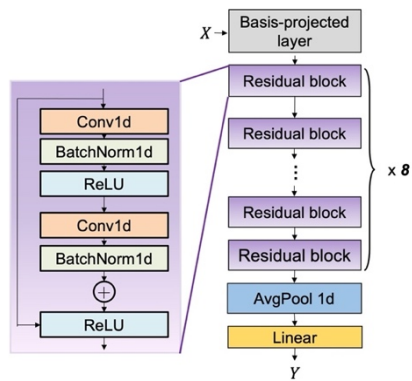

Fig. 1 DL structure of odor classifier

## RESULTS & DISCUSSION

**Model Performances of a Practical Sparse Dataset**

DL modules learned features from the non-zero terms of the input data. The top priority of handling pattern-sparse data was decreasing its high sparsity ratio. After the GC-MS spectra were fed into the BPL, the sparsity ratio of the transformed GC-MS spectra decreased from 80.38% to zero. Admittedly, a fully-connected layer (FCL) was able to achieve the same decrement for the sparsity ratio, but it did not well-performed on the pattern-sparse data (Table 1). The F1 score maintained almost the same values, around 0.56 and 0.57, whether the FCL was applied before the 1d-CNN. The models equipped with the BPL demonstrated better F1 scores than the 1d-CNN model. Moreover, the BPL-equipped models also surpassed the previously proposed model with a preprocessing of unit vectorization (Yu et al., 2021). The design of the BPL forced all of the parameters in the layer to interact with the non-zero terms in the pattern-sparse data (Eq. 1), and resulting in a better performance.

Table 1 Performance comparison of different model architectures.

| Model | F1 score | | |
|---|---|---|---|
| 1d-CNN | 0.5690 | | |
| Unit vectorization + 1d-CNN | 0.5974 | | |
| Model \ Size* | 256 | 490 | 768 |
| FCL + 1d-CNN | 0.5739 | 0.5699 | 0.5642 |
| BPL + 1d-CNN | **0.6139** | **0.6177** | **0.6344** |

*The element size of the data output after the FCL/BPL.

The number of projected bases was a hyperparameter predominantly affecting the performance of BPL-equipped models. When the bases number were 256, 490, and 768, the F1 scores of models were 0.6139, 0.6177, and 0.6344, respectively (Table 2). The model F1 scores increased when the basis number of BPL was set to a larger number. When the BPL had more learnable bases, there were greater chances to rotate these bases in refined directions. The projected bases were the implicit features learned by DL models in the training process. Namely, the projected bases could serve as a lookup table that would be referred to by the DL model in the inference process. A larger lookup table provided a higher capacity to record more data elements. Therefore, it is recommended that the number of bases would be no less than the original size of the data element. The BPL altered the dimensionality of a data element and still retained the traceability of the original data space by these projected bases.

It brought the advantage of freeing the architecture design of DL models.

Table 2 Comparison of model performance among four parameter initializing methods.

| Base Model | F1 score | | |
|---|---|---|---|
| 1d-CNN | 0.5690 | | |
| **BPL-equipped Model** | **F1 score** | | |
| Bases num. / Initializing method | 256 | 490 | 768 |
| SVD | 0.6059 (↑ 6.49%) | 0.6051 (↑ 6.34%) | - |
| NMF | 0.5950 (↑ 4.57%) | 0.5970 (↑ 4.92%) | - |
| Multivariate normal | 0.5923 (↑ 4.09%) | 0.5849 (↑ 3.59%) | 0.6192 (↑ 8.82%) |
| von Mises | **0.6139 (↑ 7.89%)** | **0.6177 (↑ 8.56%)** | **0.6344 (↑ 11.49%)** |

In order to further discuss the differences between these four initializing methods, we can trace back to the original data space. The fine-tuned projected bases were learned from the supervised DL model training process; components were factorized by unsupervised SVD and NMF. They were both data elements containing implicit features of the original data space. These bases and components were mapped onto a 2D plane via principal component analysis (Fig. 2). The SVD and NMF components revealed no noticeable scattering change after the model training procedure (Fig. 2c and 2d). The BPL-equipped models with SVD or NMF initialized bases showed 4.5-6% increments (roughly half of the von Mises sampler) on the F1 scores (Table 2). These indicated that the *N*-spheres constructed by these factorized components were not the most appropriate transformed space. The learned bases initialized by the multi-normal sampler (Fig. 2e) settled in a region close to the components factorized by SVD and NMF (Fig. 2a and 2b). The classifier even demonstrated a lower F1 score than these two unsupervised methods. As for the bases initialized by the von Mises sampler (Fig. 2f), they were located in an isolated region from other methods. Moreover, the models showed the best F1 scores. These results implied that the bases initialized by a von Mises sampler formed a unique *N*-sphere and a better manifold to represent pattern-sparse data.

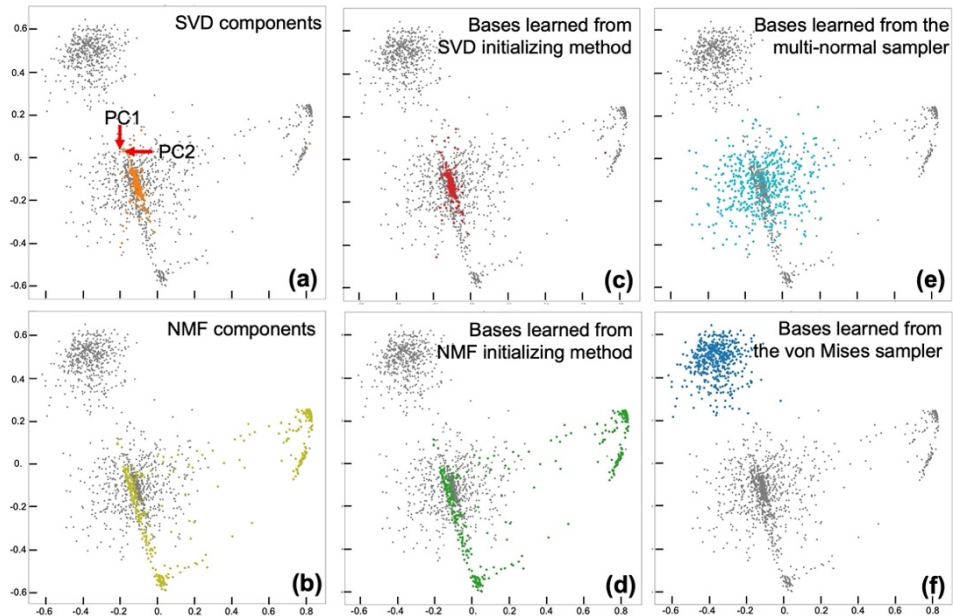

Fig. 2 Visualizations of the components calculated from matrix factorization methods and learned projected bases in BPL (*N*=490).

**Is Orthogonality Essential for Pattern-sparse Data Representation in DL Models?**

The orthogonality of the bases existed when applying matrix factorization methods, including SVD and eigen-decomposition. The orthogonal property ensured the bases fully extracted features from all dimensions. Also, all the bases were guaranteed to be linearly independent. However, due to many zero-value data, the high-sparsity data would bias the factorized components. The learned bases initiated from SVD components probably retained the orthogonal configuration. However, the DL models with BPL initialized by SVD components did not demonstrate superior performance in odor classification. The results inferred that the restriction of orthogonality did not bring advantages to reforming a pattern-sparse dataset.

## CONCLUSIONS

Sparse data heavily influenced the performance of DL models. This study proposed the basis-projected layer (BPL) to find another representation space of sparse data. It transformed the sparse data into a dense formation by projecting it onto a new *N*-sphere. A practical GC-MS dataset of coffee odor evaluated the effectiveness of BPL. The BPL-equipped model initialized by a von Mises distribution sampler showed the best performance. When the base number was 490 (same as the original data dimension) and 768, the model reached an F1 score increment of 8.56% and 11.49%, respectively. BPL preserved the pathway to trace back to the original data space and made further analysis possible. It also freed the architecture design of the DL models; therefore, pattern-sparse data could be analyzed by a more complicated model (e.g., transformer-based model) in the future.

## ACKNOWLEDGEMENTS

This work was supported by the National Science and Technology Council (NSTC), Taiwan (Grant Nos. NSTC 111-2313-B-002-056-).